\documentclass{article}
\usepackage{etex}
\usepackage[utf8]{inputenc}
\usepackage{hyperref}
\usepackage{latexsym}
\usepackage{url}
\usepackage[a4paper]{geometry}
\usepackage[pdftex,dvipsnames]{xcolor}
\usepackage{amsmath}
\usepackage{amsfonts}
\usepackage{amssymb}
\usepackage[T1]{fontenc}
\usepackage{graphicx}
\usepackage{multirow}
\usepackage{multicol}
\usepackage{xargs}
\usepackage{todonotes}
\usepackage{ctable}
\usepackage{booktabs}
\usepackage{enumerate}
\usepackage{ulem}
\usepackage{subcaption}

\title{Effective training of deep convolutional neural networks for hyperspectral image classification through artificial labeling} 
\author{\small
		Wojciech Masarczyk,
		Przemysław Głomb,
		Bartosz Grabowski,
		Mateusz Ostaszewski
}
\date{}

\newcommand{\address}{{
\footnotesize
\begin{center}
Institute of Theoretical and Applied Informatics, Polish Academy of Sciences\par\nopagebreak
Bałtycka 5, 44-100 Gliwice, Poland\par\nopagebreak
Email: \{wmasarczyk,przemg,bgrabowski,mostaszewski\}@iitis.pl\par\nopagebreak
Telephone: +48 32 2317319\par\nopagebreak
\end{center}
}}

\begin{document}

\maketitle
\address

\smallskip

\begin{abstract}
Hyperspectral imaging is a rich source of data, allowing for multitude of effective applications. However, such imaging remains challenging because of large data dimension and, typically, small pool of available training examples. While deep learning approaches have been shown to be successful in providing effective classification solutions, especially for high dimensional problems, unfortunately they work best with a lot of labelled examples available. To alleviate the second requirement for a particular dataset the transfer learning approach can be used: first the network is pre-trained on some dataset with large amount of training labels available, then the actual dataset is used to fine-tune the network. This strategy is not straightforward to apply with hyperspectral images, as it is often the case that only one particular image of some type or characteristic is available. In this paper, we propose and investigate a simple and effective strategy of transfer learning that uses unsupervised pre-training step without label information. This approach can be applied to many of the hyperspectral classification problems. Performed experiments show that it is very effective at improving the classification accuracy without being restricted to a particular image type or neural network architecture. The experiments were carried out on several deep neural network architectures and various sizes of labeled training sets. The greatest improvement in overall accuracy on the Indian Pines and Pavia University datasets is over 21 and 13 percentage points, respectively.
An additional advantage of the proposed approach is the unsupervised nature of the pre-training step, which can be done immediately after image acquisition, without the need of the potentially costly expert's time.
\end{abstract}



\section{Introduction}

Classification of hyperspectral images (HSI) has many potential applications, e.g. land cover segmentation~\cite{BioucasDias2013HSsurvey}, mineral identification~\cite{Ghamisi2017spectral}, or anomaly detection~\cite{Chang2003detclass}. The classification algorithms used include both general models, e.g. the SVM~\cite{Melgani2004svm}, and dedicated approaches, taking into account spectral properties or spatial class distribution~\cite{Romaszewski2016PNGrow}. Recently there have been attempts to use Deep Learning Neural Networks (DLNN) for HSI classification. The motivation is that such methods have gained attention after achieving state of the art in natural image processing tasks~\cite{krizhevsky2012alexnet}. Their unique ability to process an image using a hierarchical composition of simple features learned during training makes them a powerful tool in areas where manipulation of high-dimensional data is needed. 

While DLNN can achieve very good accuracy scores, they have the drawback of requiring a large amount of training data for estimation of model parameters. Such data is not always available, as it is common to have a single HSI with just a handful of training labels available. To bridge a gap between this realistic scenario and DLNN network requirements, we propose an approach that trains the DLNN in two stages, with the first -- pre-training -- stage using artificial labels. In the remainder of this section, we discuss the relevant related, and introduce the motivation of our approach and state the hypothesis that is the base of our method.

A number of DLNN architectures have been proposed, inspired by mathematical derivations and/or neuroscience studies. The Convolutional Neural Networks (CNN)~\cite{lecun1995convolutional} are a special case of deep neural networks which were originally developed to process images, but are also used for other types of data like audio. They combine traditional neural networks with biologically inspired structure into a very effective learning algorithm. They scan multidimensional input piece by piece with a convolutional window, which is a set of neurons with common weights. Convolution window processes local dependencies (features) in the input data. The output corresponding to one convolutional window is called a feature map and it can be interpreted as a map of activity of the given feature on the whole input. The CNN remain one of the most popular architectures for DLNN classification in use today.

Other approaches include the generative architectures, e.g. the Restricted Boltzmann Machine (RBM) \cite{smolensky1986information,ackley1985learning}, Autoencoders (AE)~\cite{hinton2006reducing} or Deep Belief Network (DBN)~\cite{hinton2009deep,hinton2006fast}. Yet another popular architecture is the Recurrent Neural Network (RNN) which, through directed cycles between units, has the potential of representing the state of processed sequence. They are applicable e.g. for time series prediction or outlier detection. The most popular types of RNN are Long Short-Term Memory (LSTM) networks~\cite{hochreiter1997long} and Gated Recurrent Units (GRUs)~\cite{dey2017gate}. They improve the original RNN architecture by dealing with exploding and vanishing gradient problem. 

For classification of HSI data, the CNN is the most popular architecture chosen. In~\cite{yu2017convolutional} the simple CNN architecture is adapted to HSI classification;
the lack of training labels is mitigated by adding geometric transformations to available training data points. In~\cite{lee2017going} authors use three kinds of convolutional windows: two of them are 3D convolutions which analyse spatial and spectral dependencies in the input picture, while the third is the 1D kernel. Next the feature maps from these three types of convolutions are stacked one after the other and create joint output of this first part of the network. The following layers consist only of the one dimensional convolutional kernels and residual connections. The authors of \cite{Han2018jointss}
introduce a parallel stream of processing with an original approach for spatial enhancement of hyperspectral data.
The authors of~\cite{Zhou2019limited}
design a deep network that reduces the effect of Hughes phenomenon (curse of dimensionality) and use additional unlabelled sample pool to improve performance.
In~\cite{Xu2018rpnet}
authors propose an alternative architecture called RPNet based on prefixed convolutional kernels. It combines shallow and deep features for classification. Another architecture (MugNet) is proposed in~\cite{Pan2018mugnet}
with a focus on simplicity of processing for classification of hyperspectral data with few training samples and reduced number of hyperparameters. A yet another architecture approach is used in~\cite{Gao2019multibranch}
where a multi-branch fusion network is introduced, which uses merging multiple branches on an ordinary CNN. An additional L2 regularization step is introduced to improve the generalization ability with limited number of training samples. The work \cite{Zhao2019layers}
proposes a strategy based on multiple convolutional layers fusion. Two distinct networks, composed of similar modules but different organization, are examined. 

Other architectures are also used. For example in~\cite{mou2017deep} authors utilize the sequential nature of hyperspectral pixels and use some variations of recurrent neural networks -- Gated Recurrent Unit (GRU) and Long-Short Term Memory (LSTM) networks. Moreover, in~\cite{wu2017convolutional} one dimensional convolutional layers followed by LSTM units were used. Chen et. al.~\cite{chen2014deep} use artificial neural networks for feature extraction. They utilize stacked autoencoders (SAE) for feature extraction from pixels, and PCA for reduction of the spectral dimensionality of the training segments taken from the picture. Next, the logistic regression is performed on this spectral (SAE) and spatial (PCA) extracted information. Another approach~\cite{Fan2019infestation}
uses stacked SAE for an application study -- detection of a rice eating insect. RNN architectures are also employed, as they are suitable for processing the spectral vector data. The work \cite{Guo2018guidedrnn}
applies sequential spectral processing of hyperspectral data, using a RNN supported by a guided filter. In \cite{Shi2018multirnn} authors
use the multi-scale hierarchical recurrent neural networks (MHRNNs) to learn the spatial dependency of non-adjacent image patches in the two-dimension (2D) spatial domain. Another idea to analysing HSI is spatial--spectral method in which network takes information not only from spectrum bands but also from spatial dependencies of image~\cite{lee2017going}.

A significant problem in practical hyperspectral classification is the small number of training samples. It is related to the difficulty of obtaining verified labels~\cite{BioucasDias2013HSsurvey}, as often each pixel must be individually evaluated before labelling. Therefore, a reference hyperspectral classification experiment may assume number as low as 1\% available samples per class~\cite{Ghamisi2017spectral}. A number of approaches has been exploited to deal with this difficulty, e.g. including combining spatial and spectral features~\cite{Plaza2009recent}, additional training sample generation~\cite{Cholewa2019lvariant}, extending the classification algorithm with segmentation~\cite{tarabalka2010segmentation}, or employing Active Learning~\cite{dopido2013semi}. 

For the DLNN classification, the lack of high volume of training data is a serious complication, as they typically require a lot of data to achieve high efficiency. Optimal use of DLNN in HSI classification would require learning them with just a few labelled samples. This may be obtained by searching for well-tailored architecture for specific task~\cite{yu2017convolutional}, however such approach requires relatively big validation set to obtain meaningful results. The other approach is to expand the available training set. It may be achieved either by artificially augmenting training set or using different dataset as a source for pre-training~\cite{wu2018finetuning}. Another approach is to add a regularization step to improve the generalization ability with limited number of training samples~\cite{Gao2019multibranch}. A simplification of the network architecture for classification with few training samples is employed in the MugNet network~\cite{Pan2018mugnet}. Finally, where possible, the transfer learning approach is used, e.g.~\cite{windrim2018transfer5}.

The transfer learning~\cite{pan2010transfer0} uses training samples from two domains, 
which share common characteristics. A network is first pre-trained on the first domain, 
which has plentiful supply of training samples but does not solve the problem at hand. 
Then, the training is updated with the second domain, which adapts the weights to 
the actual problem. 

Transfer learning is simple to apply in the case of convolutional neural networks (CNNs). In \cite{yosinski2014transfer}, authors compared different versions of transfer learning for CNNs in the case of natural images classification. They studied its effects depending on the number of transferred layers and whether they were fine tuned or not as well as depending on the differences between the considered datasets. In \cite{ng2015transfer2}, authors used transfer learning on CNNs to recognize emotions from the pictures of faces. Other uses include evaluating the level of poverty in a region given its remote sensing images \cite{xie2015transfer3} and computer-aided detection using CT scans \cite{shin2016transfer4}.

There have been applications of transfer learning in the general remote sensing (not-hyperspectral) images. In~\cite{Zhou2016tlfeatures}
deep learned features are transferred for effective target detection; negative bootstrapping is used for improving the convergence of the detector. A similar approach is applied in~\cite{Lyu2017cities} 
where RNN network trained on multispectral city images is used to derive features for studying urban dynamics across seasonal, spatial and annual variance. The authors of~\cite{Hu2015hires}
study the performance of transfer learning in two remote sensing scene classifications. The results show that features generalize well to high resolution remote sensing images. As the work~\cite{Lyu2016change}
shows, transfer learning can be applied in remote sensing using RNN architectures also.

Recently, transfer learning has been also applied to the HSI data. In \cite{windrim2018transfer5}, authors applied it for CNNs originally used for classifying well known remote sensing hyperspectral images to classify images acquired from field-based platforms and regarding a different domain. The authors of~\cite{Li2017tlad}
use a intermediate step of supervised similarity learning for anomaly detection in unlabelled hyperspectral image. A different approach to transfer learning is proposed in~\cite{Lin2018deep}
which explores the high level feature correlation of two HSI. A new training principle simultaneously processes both images, to estimate a common feature space for both images. A yet another approach is proposed in~\cite{Yuan2017superres}
where HSI superresolution is achieved using supported high resolution natural image. This natural image is used as a training reference, which is later adapted to HSI domain. 
In~\cite{Fang2018dual},
iterative process combines training and updating the currently used training label set. Two specialized architectures (for spatial and spectral processing) are used. The training iteratively extends the current label set, starting from the initial expert's labels. 

\label{utla}

The above approaches do not apply to the arguably most popular practical scenario, where only a single HSI with a handful of labels is available. Moreover, getting the training labels often requires additional resources (e.g. expert consultation and/or site visit). It is thus desirable to have unsupervised methods for realization of the pre-training step. Authors of \cite{Du2013unsuptl} 
use outlier detection and segmentation to provide candidates for training of target detector in HSI. This information is used to construct a subspace for target detection by transfer learning theory. This shows the potential of using an unsupervised approach, however limited to separation of target/anomaly points from the background. In the work of~\cite{wu2018finetuning},
a separate clustering step is used for generation of pseudo-labels, using Dirichlet process mixture model. The network is trained on the pseudo-labels, then the all but last layers are extracted, and the final network is trained on the originally provided training labels. While this scheme is shown to be effective in the presented results, it relies on a complex non-neural preprocessing and tailoring the DLNN configuration to each dataset separately. Also, the effect of size of label areas and effects on different architectures are not investigated. We show that similar gains can be made with a simpler preprocessing, independent of the DLNN architecture chosen. The authors of~\cite{Raina2007self}
propose to use a sparse coding to estimate high level features from unlabelled data from different sources. This approach does not require training data, but is tailored to the case where multiple inputs are available, preferably with diverse contents.

To close the gap between data inefficient deep learning models and practical applications of HSI we propose a method which takes advantage of abundant unlabelled data points present on HSI images. Precisely, we state a hypothesis: \emph{Spatial similarity of unlabelled data points can be utilized to gain accuracy in hyperspectral classification.}
To corroborate our hypothesis, we construct a simple clustering method that assigns artificial label to each pixel on the image based on its spatial location. This artificial dataset is used to pre-train deep learning classifier. Next the model is fine-tuned with original dataset. Through series of experiments we show superiority of the proposed approach over the standard learning procedure. 
Our approach is motivated by two known phenomena: cluster assumption~\cite{chapelle2006} and regularization effect of noise in classes~\cite{rolnick2018labelnoise, hinton2015distilling, salimans2016improved}. We note that many of remote sensing
images share common properties, most notably the `cluster assumption' -- 
pixels that are close to one another or form a distinct cluster or group 
frequently share the class label. Additionally, due to the simplistic form of our clustering method, we purposefully introduce noise in labels used during pre-training phase, however as shown in~\cite{rolnick2018labelnoise} this label noise has little to no effect on final accuracy, as long as number of properly labelled examples scales proportionally which is our case.

\section{Materials \& Methods}

Our method is to be applied in the following case:
\begin{enumerate}
	\item Classification of pixels from a remote sensing hyperspectral image;
	\item Neural networks used as a classifier;
	\item Few training labels available.
\end{enumerate}
In such situation, we propose to augment the training with a pre-training step that uses artificial labels, which are independent of the training labels. Inclusion of this pre-training step can be viewed as a modification of a transfer learning approach. Conventional transfer learning in this case would use a related dataset (source domain) with abundance of labels to pre-train, then the current dataset (target domain) to fine-tune. In our case, the source domain consists of every point in the hyperspectral image, while the target domain is composed of only the labelled samples. In the remainder of this Section we discuss: the spatial structure of hyperspectral images and the characteristics of neural network that make this approach feasible, and the details of its application. We also describe the experiments used to test the proposed approach.

\subsection{Spatial structure of hyperspectral images}

It is well-known that remote sensing hyperspectral images contain spatial structure, that can be exploited to improve classification scores when only a few training samples are available~\cite{tan2014efficient,tarabalka2010segmentation,Romaszewski2016PNGrow,wang2014semi,Cholewa2019lvariant}. A segmentation can be applied to propose candidate pixels for labelling with high confidence~\cite{tan2014efficient} or identify connected components for label assignment~\cite{tarabalka2010segmentation}. Class training samples can be extended through moderated region growing~\cite{Romaszewski2016PNGrow} or spatial filtering combined with spatial-spectral Label Propagation~\cite{wang2014semi}. Finally, disagreement between spatial and spectral classifiers can be used to propose new samples~\cite{Cholewa2019lvariant}. A qualitative investigation of this phenomenon shows that hyperspectral pixels close to one another, whether spatially or spectrally, are likely to have the same class label, thus fulfilling the `cluster assumption'~\cite{chapelle2006}. This effect often leads to a blob-like structure of a hyperspectral dataset, observed in many hyperspectral classification problems (e.g. land cover labelling in remote sensing, paint identification in heritage science, scene analysis in forensics). A single class with samples in different parts of an image can be made of a number of blobs, which differ from each other because of, e.g., non-uniformity in class structure (e.g. the same class can contain differing crop types), spectral variations (e.g. same crop in two areas can have differing properties due to sunlight exposure, soil type) or acquisition conditions (e.g. level of lighting, shadows). 

\subsection{Emergence of data-dependent filters in neural network training}\label{representation}

During training, subsequent layers of a deep neural network form a representation of a local input data structure~\cite{Goodfellow2016deep}. Given a data source, this representation, especially on lower layers, can be remarkably similar across different dataset. For example, in the problem of natural image classification the learned kernels resemble a Gabor filter bank~\cite{krizhevsky2012alexnet,Lee2009ConvolutionalHierarchical}, independent of class set. This form of a filter can be shown to arise independently when independent components~\cite{Bell1997ICA} or an effective sparse code~\cite{Olshausen1996} for natural images is estimated. Another case where data-dependent filters emerge is the pretext task approach, e.g. \cite{Dai2015pretext,Howard2018Universal}, where the network first learns to predict the input sequence without class labels, which are introduced at a fine-tune stage for to get the final classification model. Apparently the deep neural networks are able, at least in part, to extract an efficient class-independent data representation. This phenomenon has not been studied for hyperspectral images, however, it can be argued that similar class-independent but data-dependent representation is being learned in training for hyperspectral image classification.

\subsection{Methods used for proposed artificial labelling approach}\label{artificial-labeling}

\begin{figure}[t]
	\begin{center}
		\includegraphics[width=\textwidth]{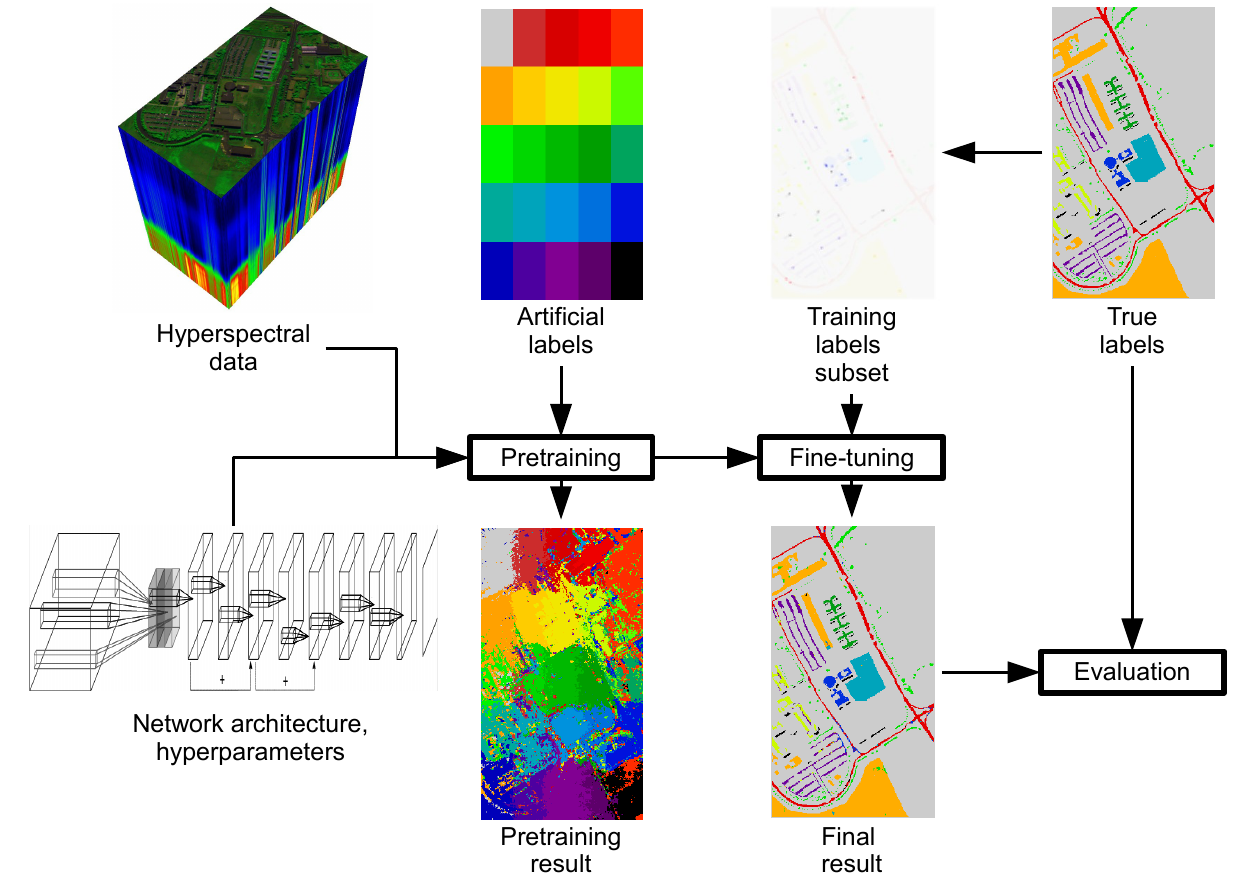}
		\caption{The overview of the unsupervised pretraining algorithm proposed in this work. First, the network is pretrained on grid-based scheme of artificially assigned labels. The network weights are then fine-tuned on a limited set of training samples selected from true labels, consistent with typical hyperspectral classification scenario.}
		\label{fig:labels}
	\end{center}
\end{figure}

Our method for creating artificial labels for the pre-training step is a simple segmentation algorithm which assumes the local homogeneity of samples' spectral characteristics. It works by dividing the considered image into $k$ rectangles, where each of these rectangles has its own label. For an image of height $h$ and width $w$, we divide its height into $m$ roughly equal parts and its width into $n$ roughly equal parts, so that $k=m \cdot n$. We then get $k$ rectangles, where each one's height equals approximately $h/m$, while its width equals approximately $w/n$. Each of these rectangles defines a different artificial class with a different label. A schematic is presented in Figure~\ref{fig:labels}.

The function of artificial labels is for the network to learn class-independent blob patterns present in the data. This focuses the network training in the fine tuning on the actual training labels, with the network `oriented' towards the features of the current image. It can also be of advantage in situations when a class is composed of multiple blobs, and not all of them have samples in the training set. In that case sufficiently correct labelling is unlikely to be obtained~\cite{wang2010new} with just the training samples, but the proposed grid structure forces the network to estimate features for the whole image. An additional advantage of this approach is to shift the potentially time consuming pre-training from the expert labelling moment to the acquisition moment. In other words, network training does not need to be held back until the expert's labels are available, but can commence right after the image is recorded.

\subsection{Selected network architectures}

In our experiments three architectures were tested, based on~\cite{lee2017going,yu2017convolutional,liu2018siamese}. All three share a common approach to exploit local homogeneity of hyperspectral 
images, however each one has its unique strengths and weaknesses making 
them an interesting testbed for the universality of the proposed method.
The first architecture~\cite{lee2017going} features relatively high number of convolutional 
layers which might be helpful in transfer learning application. 
The second architecture~\cite{yu2017convolutional}, to the best of authors knowledge, is one of the 
best networks that are trained on limited number of samples per class. However due to its constrained capacity, it may not benefit as much from the pre-training phase.
The last of the considered convolutional neural networks~\cite{liu2018siamese} is conceptually the
simplest of the three, which allows us to test our approach using more
conventional convolutional architecture.

\subsection{Experiments}
\label{exp-desc}

This subsection describes the experiments evaluating the proposed approach. We investigate the performance of the artificial label pre-training in the following four experiments:
\begin{enumerate}
	\item Experiment 1 evaluates the accuracy improvement achieved by using the method.
	\item Experiment 2 investigates the variability introduced by the size and shape of the patches used to define artificial classes, using one of the datasets from Experiment 1.
	\item Experiment 3 is an additional investigation of the observed phenomenon that  big patches (larger area, smaller total number of classes) perform worse than little ones (smaller area, larger total number of classes), done using a different dataset.
	\item Experiment 4 is an examination of a claim about the emergence of data-independent representations during neural network training using proposed artificial labelling scheme.
\end{enumerate}
In the following subsections, the detailed descriptions of the conducted experiments are given, while in the Section~\ref{sec:results} we present the results of the experiments.

\subsubsection{Experiment 1}\label{exp1-description}

In this experiment, the proposed approach is evaluated using different hyperspectral images and neural network architectures to prove its robustness.

For the experiment, we have used two well-known hyperspectral datasets: Indian Pines and Pavia University.

The Indian Pines dataset was collected by the AVIRIS sensor over the Northwest Indiana area. The image consists of $145 \times 145$ pixels. Each pixel has $220$ spectral bands in the frequency range 0.4--2.5 $\times10^{-6}$ m. Channels affected by noise and/or water absorption were removed (i.e. [104--108], [150--163], 220), bringing the total image dimension to $200$ bands. The reference ground truth contains $16$ classes representing mostly different types of crops. To be consistent with experiments performed in~\cite{lee2017going}, we choose only $8$ classes.

The Pavia University dataset was collected by the ROSIS sensor over the urban area of the University of Pavia in Italy. This image consist of $610\times 340$ pixels. It has $115$ spectral bands in the frequency range from 0.43 to 0.86 $\times10^{-6}$ m. The noisiest 12 bands were removed, and remained 103 were utilized in the experiments. Ground truth includes $9$ classes, corresponding mostly to different building materials. 

The two datasets were subjected to a feature transformation. For a given dataset, the mean $m_{b}$ of each hyperspectral band $b$ were calculated. In the case of each dataset, and for each given pixel $x$ and band $b$, the corresponding mean $m_{b}$ was subtracted, $x(b) := x(b) - m_{b}$.

For this experiment, all three of the previously introduced neural network architectures were used. As discussed previously, the training was divided into pre-training and fine-tuning stages. In pre-training, the data was labelled through assigning an artificial class to each block within a grid of dimensions $5\times 5$. No ground truth data was used at this stage. In the fine-tuning stage, a selected number of ground truth labels was used. The number of training samples from each class was set at $n=5,15,50$. This allowed to observe the performance both in typical hyperspectral scenarios (small number of classes used) and deep network scenarios (larger number of samples per class available). Because the classification accuracy depends on the training set used in fine-tuning each experiment was repeated $n=15$ times for error reporting. 
The performance is reported in Overall Accuracy (OA) after fine tuning. Additionally, Average Accuracy (AA) and $\kappa$ coefficient were inspected and improvements verified with statistical tests.

\subsubsection{Experiment 2}\label{exp2-description}

The second experiment investigates the variability introduced by the size and shape of the patches used in artificial labelling.

For this experiment only the Indian Pines image introduced in Experiment 1 was used, as it is the more challenging of the two introduced datasets. As was the case with the previous experiment, the mean was subtracted. The network investigated is the architecture based on~\cite{lee2017going}, chosen because it has the most potential to be affected by the transfer learning process.

\begin{figure}[t]
	\centering
	\begin{subfigure}[t]{0.35\textwidth}
		\includegraphics[width=\linewidth]{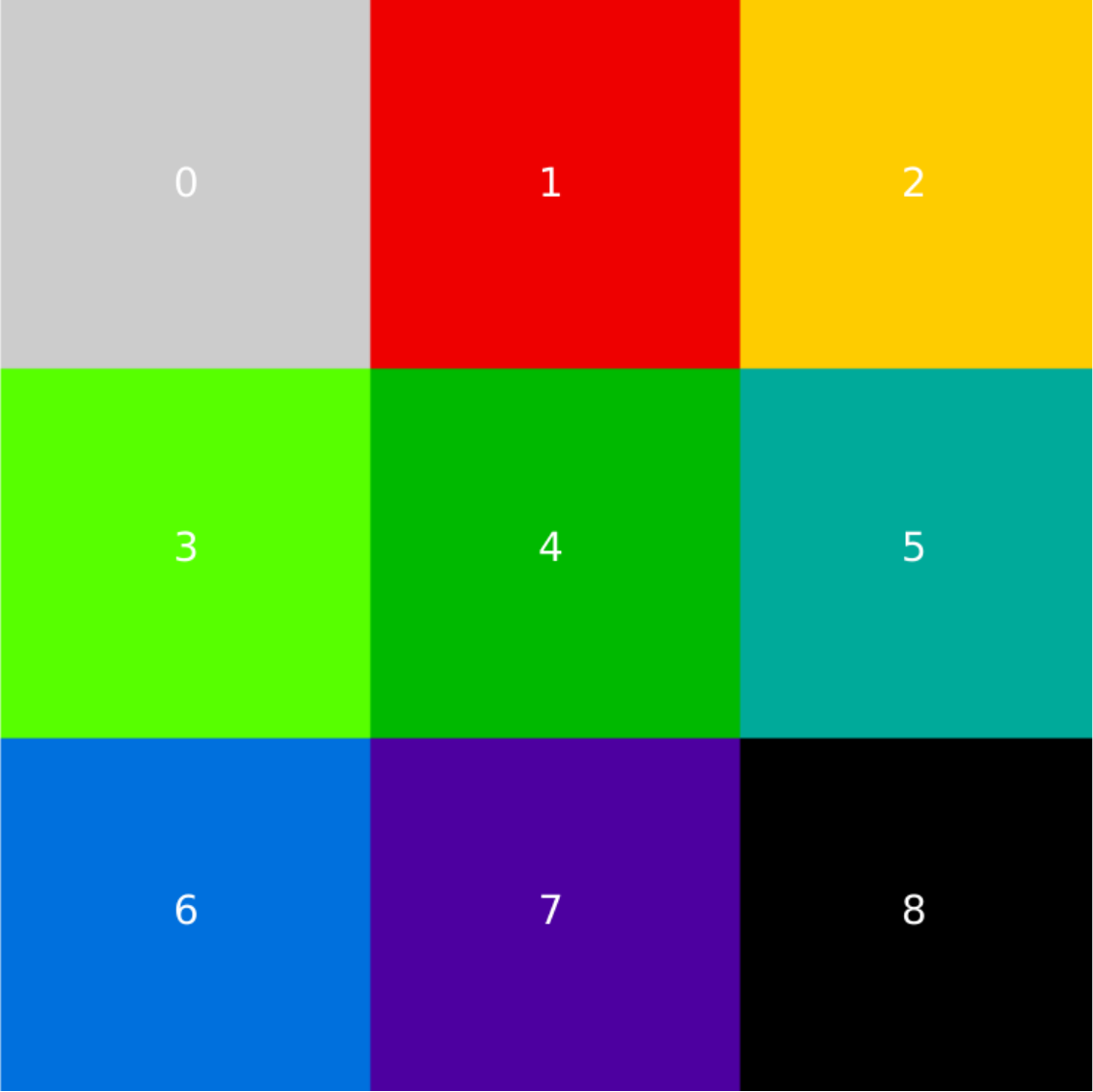}
		\caption{Grid $3\times 3$}
	\end{subfigure}\hspace*{0.1\textwidth}
	\begin{subfigure}[t]{0.35\textwidth}
		\includegraphics[width=\linewidth]{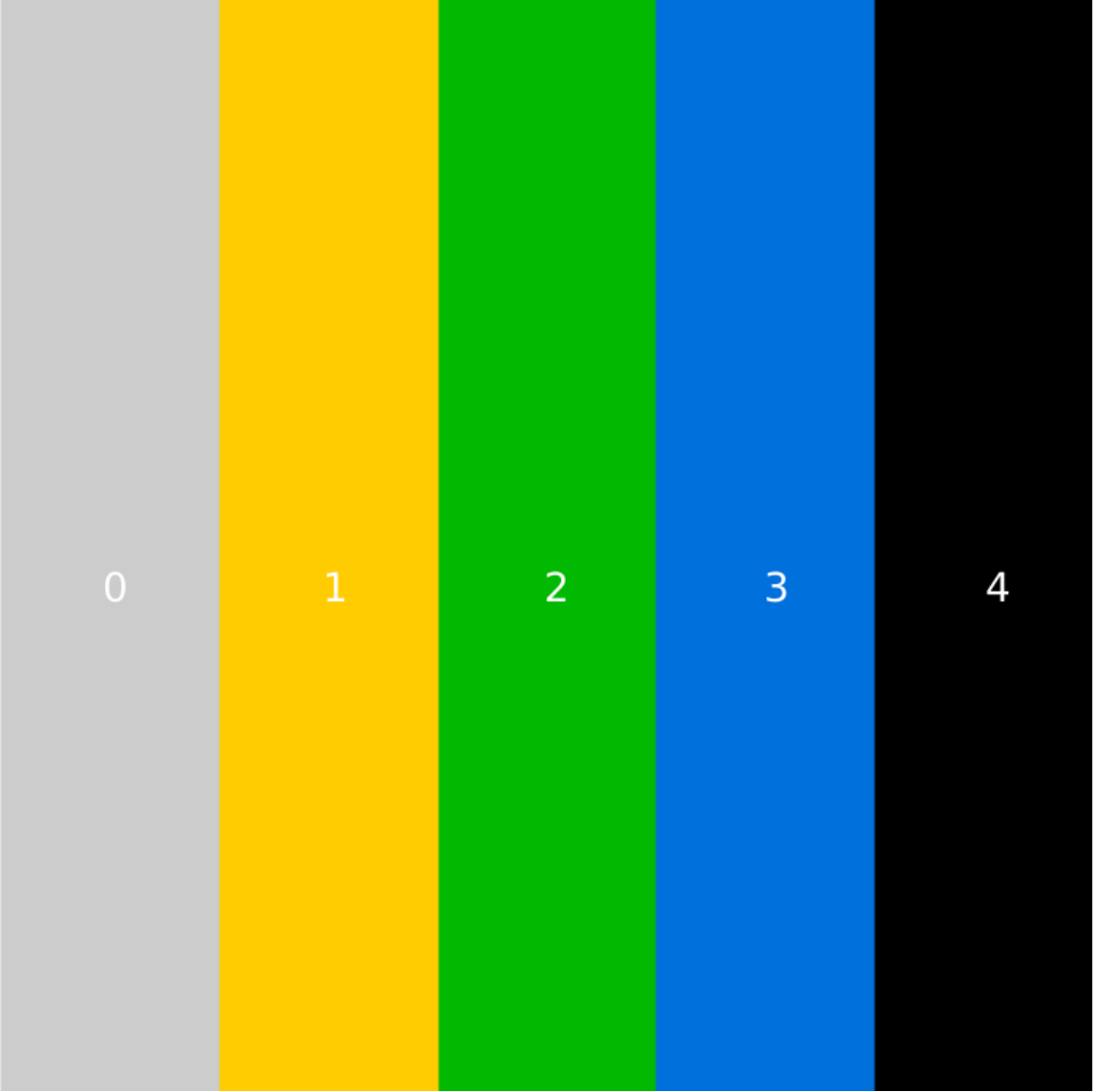}\hspace*{0.02\textwidth}
		\caption{Stripes 5}
	\end{subfigure}
	\caption{Scheme of creating artificial classes on Indian Pines dataset. From left: grid of 9 artificial classes \textbf{(a)}, vertical stripes with 5 artificial classes \textbf{(b)}. Artificial classes for Pavia University dataset were created analogically.}
	\label{fig:grids_n_stripes}
\end{figure}

In this experiment, first the grid size was investigated. The dimensions of the patches, varies from $2 \times 2$, which equals $4$ artificial classes, up to $72 \times 72$, $5184$ artificial classes. Furthermore, another way of creating artificial labels is considered. The image is divided into the given number of vertical stripes. Visualisation of different artificial labelings is presented in Figure~\ref{fig:grids_n_stripes}.

The investigated patches were created by dividing horizontal and vertical side of an image into $w=2,3,5,7,9,15,19,25,29,36,39,48,72$ equal parts. The vertical stripes were created by dividing horizontal side of an image into $s=2,5,9,16,25,36,49,81$ equal parts (so in the case of $s=2$, there are only $2$ classes located to the right and left of a single vertical line). The vertical stripes were included to observe whether the pixel distance affects the performance -- for patches, all the pixels share similar neighbourhood; for stripes, the top and bottom pixels have a notable spatial separation and, arguably, the distant pixels should not be marked with the same class label without prior knowledge of spatial class distribution. Note that in case of patches made by dividing each side of an image into $w = 29, 36, 39, 48, 72$ equal parts the size of a square patch is smaller than the size of a processing window $5 \times 5$ in tested architecture. That means no sample fed to a network during pre-training phase has a coherent class representation (i.e. a single class present in the window). This experiment was performed with $5$ training samples per class and $50$ experiment runs for each grid density and the number of stripes.

\subsubsection{Experiment 3}\label{exp3-description}

In this experiment, we test the hypothesis that the more numerous patches' division produces a better pre-training set than the less numerous ones. We investigate this using a specially designed hyperspectral test image.

In this experiment, we use the image of paints from museum's collection. This dataset~\cite{grabowski2018automatic} was collected by the SPECIM hyperspectral system in the Laboratory of Analysis and Nondestructive Investigation of Heritage Objects (LANBOZ) in National Museum in Kraków. This image consist of $455\times 310$ pixels. Each pixel has 256 spectral bands in the frequency range from 1000 to 2500 nm. Ground truth consists of manual annotations of different green pigments used in the mixture of paints for various painting regions. The image of oil paints on paper was used, selected from four available, as it was considered one of the more challenging of the images.

The layout of classes present in this image was especially designed to verify hyperspectral classifiers. The different chemical compositions of the paints used introduce variations of class spectra, yet at the same time all paints are variations of the green pigment with more or less greenish hue. The classification problem is thus difficult, but not exceedingly so. Regular grid layout, with different thickness of paints and fragments where one pigment overpaints another, introduce spatial diversity in the spectra. Since the image is artificially created, ground truth can be precisely marked. The original purpose of the image was to evaluate identification of copper pigments, difficult to differentiate by other (non-hyperspectral) sensors. Here we take advantage of its regularity by complementing the original ground truth ($n_{GT}=5$ classes) with a joined set ($n_{GT-2}=2$ classes) and split set  ($n_{GT-10}=10$ classes). Those two sets of modified ground truth allow us to compare the proposed grid scheme, as tested in experiments 1 and 2, with a ground truth based pre-training with more and less classes than the original set. We argue that the regular layout of this image is more suited for this experiment than e.g. Indian Pines or Pavia University images; usage of additional dataset allows us to further verify the generalization potential of our approach.

In the case of this dataset, the mean was subtracted as in the case of the previous experiments. Additionally, the standard deviation $\sigma_{b}$ of each hyperspectral band $b$ was calculated and then all pixels were divided by the corresponding standard deviation value $\sigma_{b}$, $x(b) := \frac{x(b)}{\sigma_{b}}$. In this experiment, as in the previous one, the neural network based on~\cite{lee2017going} was used. Training size was equal to 5 training samples per class and there were 50 experiment runs for each examined case.

The following cases were investigated: 
\begin{enumerate}
\item The performance of DLNN with pre-training performed with 2 classes prepared from joining the ground truth classes (GT-2).
\item The performance of DLNN with pre-training performed with 10 classes prepared by splitting the ground truth classes (GT-10).
\item The performance of DLNN without pre-training (GT).
\item The performance of DLNN with pre-training with artificial patches of size $5\times 5, 20\times 20, 30\times 30$.
\end{enumerate}

\begin{figure}[t]
	\centering
	\begin{subfigure}[t]{0.2\textwidth}
		\includegraphics[width=0.91\linewidth]{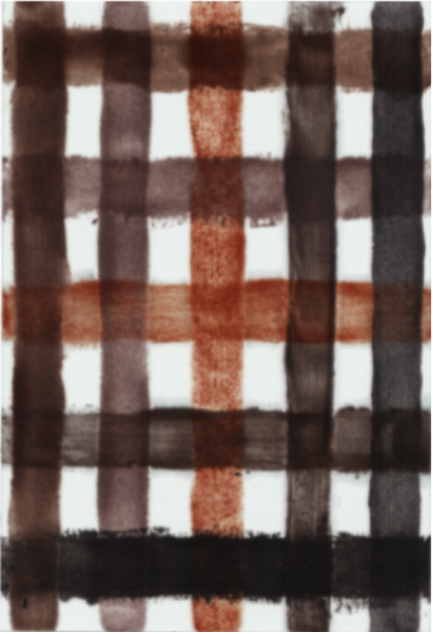}\hspace*{0.01\textwidth}
		\caption{Painting}
		\label{fig:setup:painting}
	\end{subfigure}
	\begin{subfigure}[t]{0.2\textwidth}
		\includegraphics[width=0.9\linewidth]{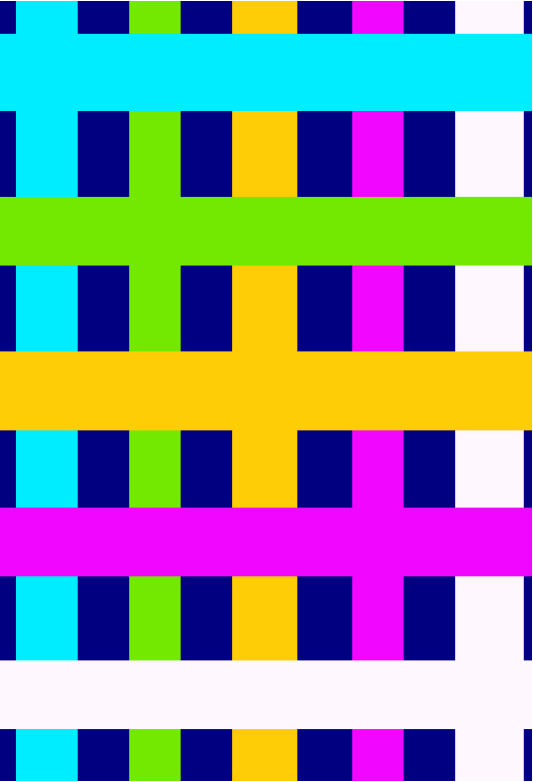}\hspace*{0.02\textwidth}
		\caption{GT}
		\label{fig:setup:gt}
	\end{subfigure}
	\begin{subfigure}[t]{0.2\textwidth}
		\includegraphics[width=0.9\linewidth]{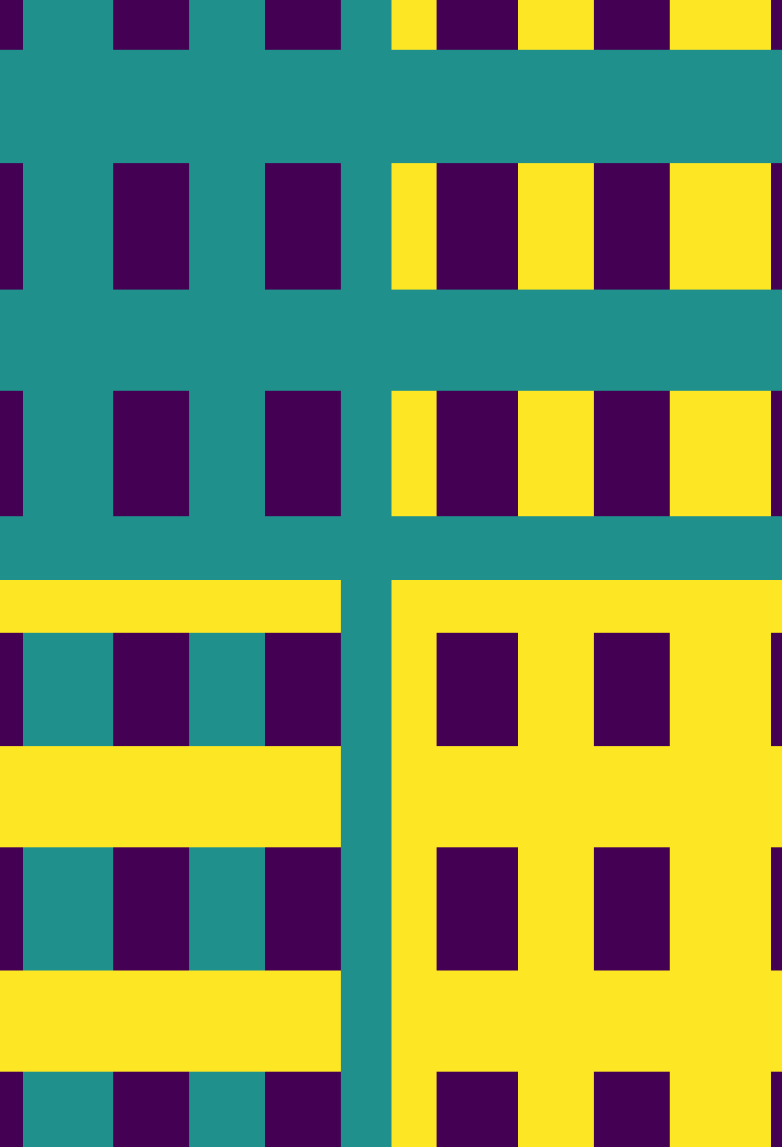}\hspace*{0.02\textwidth}
		\caption{GT-2}
		\label{fig:setup:gt2}
	\end{subfigure}
	\begin{subfigure}[t]{0.2\textwidth}
		\includegraphics[width=0.9\linewidth]{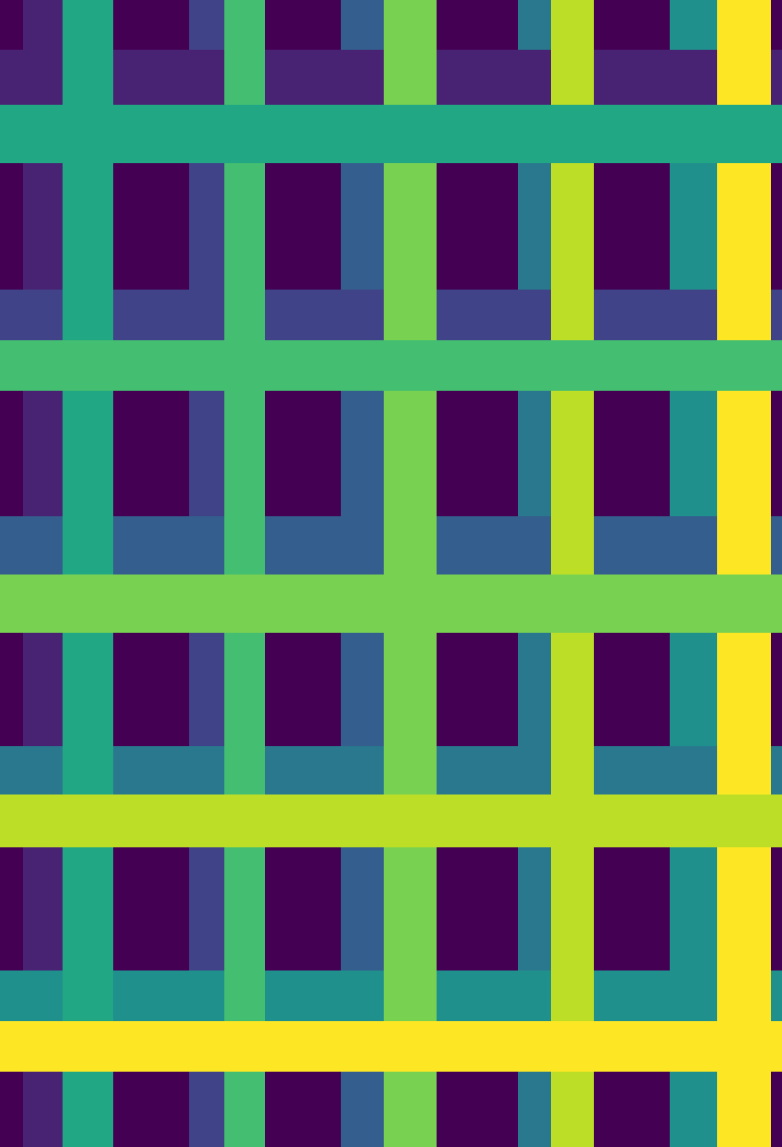}
		\caption{GT-10}
		\label{fig:setup:gt10}
	\end{subfigure}
	\caption{Scheme of creating artificial classes on Pigment dataset. From left: false-colour RGB (bands 50, 27, 17) image of the painting \textbf{(a)}, original class labels \textbf{(b)}, classes artificially joined into 2 sets \textbf{(c)}, classes artificially split into 10 sets \textbf{(d)}. Dark rectangles denote background, excluded from the experiment.}
	\label{fig:setup}
\end{figure}

\subsubsection{Experiment 4}\label{tsne-desc}

In this experiment, we examine the claims from subsection~\ref{representation} about the emergence of data-depenedent representations during neural network training using proposed artificial labelling scheme with noisy labels. To this end, we visualised internal network parameters resulting from network training using t-SNE algorithm~\cite{tsne}. In the experiment, we used neural network architecture based on~\cite{lee2017going} and the Indian Pines dataset described in subsection~\ref{exp1-description}. We trained the network on the dataset using the following scenarios:
\begin{enumerate}
\item The network was trained using $1600$ labelled samples, with $200$ samples per class. This scenario represents the neural network trained with abundant information about the data -- unrealistic, but convenient from the point of network's requirements.\label{tsne-many}
\item The network was trained using $40$ labelled samples, with $5$ samples per class. This scenario represents the neural network trained with very limited information about the data -- realistic, but difficult learning problem.\label{tsne-few}
\item The network was trained using only the artificial labels created as explained in subsection~\ref{artificial-labeling}. Therefore, the network did not 'see' the true labels and could create the internal represenations only based on the noisy labels provided for training.\label{tsne-art}
\item The network was trained using the complete pretraining-finetuning scheme introduced in this section. That is, first it was pretrained using artificial labels as in point~\ref{tsne-art}, and then all layers except the last was finetune using the training set analogous to the one from point~\ref{tsne-few}. This scenario was introduced to help explain the impact of the finetuning step in our approach.\label{tsne-ft}
\end{enumerate}

As a result, we obtain $4$ trained neural networks. As a next step, using validation dataset we extract the activations of the next-to-last layers of the considered networks, and use t-SNE algorithm, which is used to visualise high-dimensional data, to learn if the layers right before the classification layers of the networks did learn useful data representations.

\section{Results}

\label{sec:results}

This Section presents the results of experiments introduced in subsection~\ref{exp-desc}.

\subsection{Experiment 1}

The first experiment's results are presented in Table~\ref{tab:exp1}. Each column presents the result for one type of network, each row for a set dataset and the number of training examples. Each table cell presents the results with and without pre-training, in percent of Overall Accuracy, including the standard deviation of the result. The results from Table~\ref{tab:exp1} were computed from a batch of $n=15$ independent runs for each case. The specific value of $n$ was chosen to provide robust result, after a set of preliminary runs with different $n$ values. A Mann--Whitney \emph{U} test was performed on the results to confirm statistical significance of the improvement gained with the proposed method. As Overall Accuracy can be sensitive to class imbalances, Average Accuracy and $\kappa$ coefficient were computed for additional verification, and were inspected for negative performance.

\begin{figure*}[!h]
  \centering
  \begin{subfigure}[t]{0.32\textwidth}
    \centering
    \includegraphics[width=\linewidth]{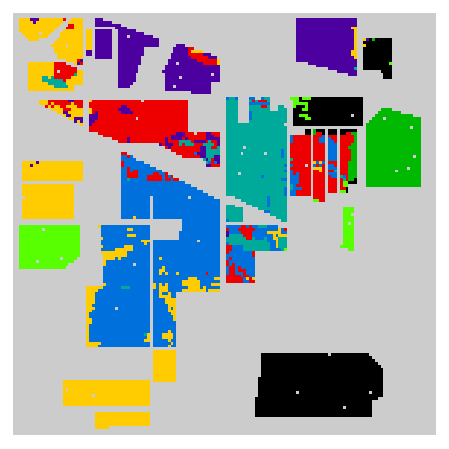}
    \caption{{\footnotesize 5s/A9 OA $76.1$\% AA $79.5$\% $\kappa$ $0.71$}}
  \end{subfigure}\hspace*{0.02\textwidth}
  \begin{subfigure}[t]{0.32\textwidth}
    \centering
    \includegraphics[width=\linewidth]{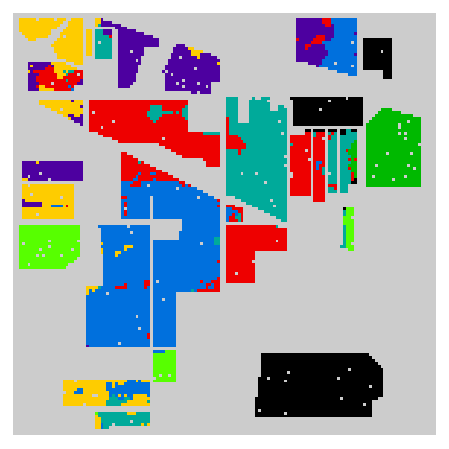}
    \caption{{\footnotesize 15s/A9 OA $88.3$\% AA $91.1$\% $\kappa$ $0.86$}}
  \end{subfigure}\hspace*{0.02\textwidth}
  \begin{subfigure}[t]{0.32\textwidth}
    \centering
    \includegraphics[width=\linewidth]{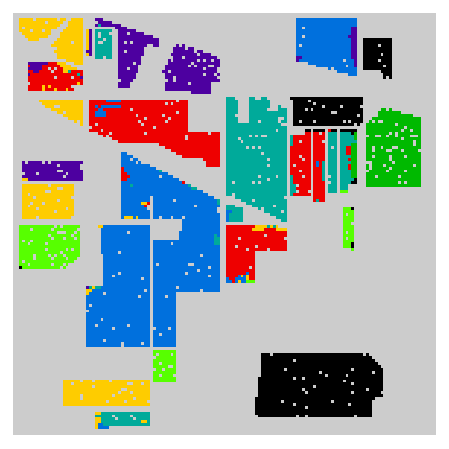}
    \caption{{\footnotesize 50s/A9 OA $94.2$\% AA $96.0$\% $\kappa$ $0.93$}}
  \end{subfigure}\\\vspace{2mm}
  \begin{subfigure}[t]{0.32\textwidth}
    \centering
    \includegraphics[width=\linewidth]{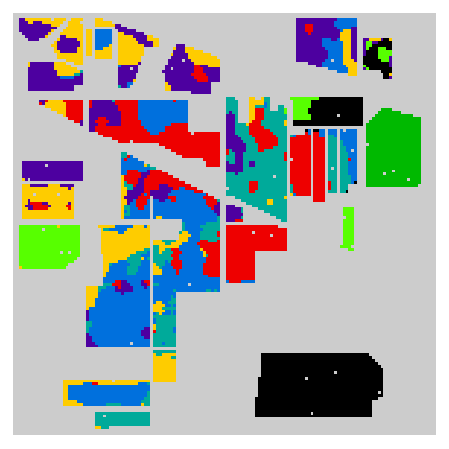}
    \caption{{\footnotesize 5s/A3 OA $66.3$\% AA $71.2$\% $\kappa$ $0.61$}}
  \end{subfigure}\hspace*{0.02\textwidth}
  \begin{subfigure}[t]{0.32\textwidth}
    \centering
    \includegraphics[width=\linewidth]{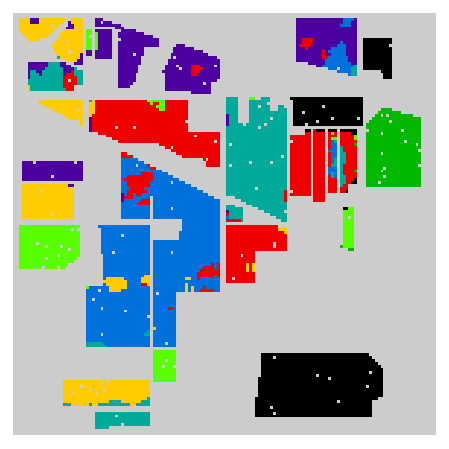}
    \caption{{\footnotesize 15s/A3 OA $87.0$\% AA $91.5$\% $\kappa$ $0.84$}}
  \end{subfigure}\hspace*{0.02\textwidth}
  \begin{subfigure}[t]{0.32\textwidth}
    \centering
    \includegraphics[width=\linewidth]{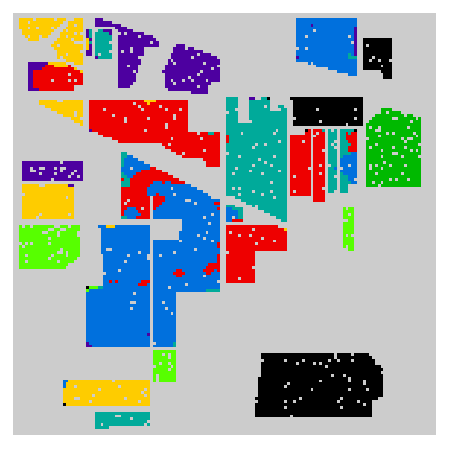}
    \caption{{\footnotesize 50s/A3 OA $94.8$\% AA $96.9$\% $\kappa$ $0.94$}}
  \end{subfigure}\\\vspace{2mm}
  \begin{subfigure}[t]{0.32\textwidth}
    \centering
    \includegraphics[width=\linewidth]{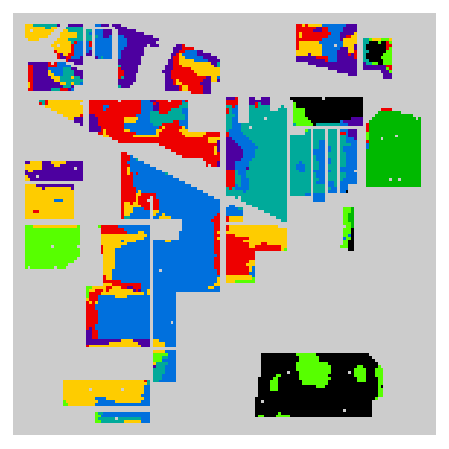}
    \caption{{\footnotesize 5s/A5 OA $65.8$\% AA $68.9$\% $\kappa$ $0.59$}}
  \end{subfigure}\hspace*{0.02\textwidth}
  \begin{subfigure}[t]{0.32\textwidth}
    \centering
    \includegraphics[width=\linewidth]{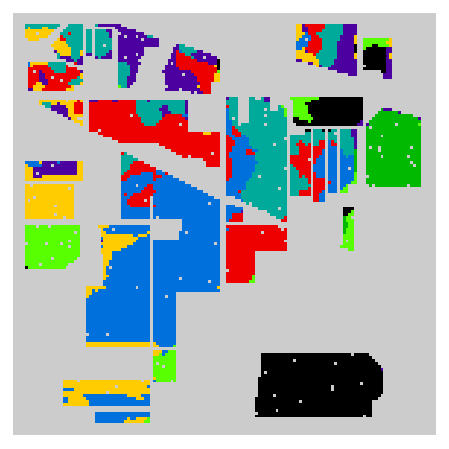}
    \caption{{\footnotesize 15s/A5 OA $75.4$\% AA $76.7$\% $\kappa$ $0.71$}}
  \end{subfigure}\hspace*{0.02\textwidth}
  \begin{subfigure}[t]{0.32\textwidth}
    \centering
    \includegraphics[width=\linewidth]{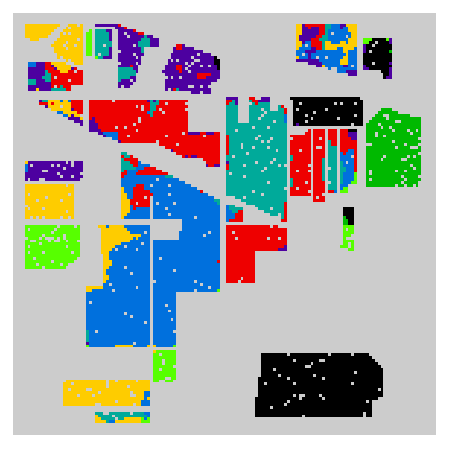}
    \caption{{\footnotesize 50s/A5 OA $87.3$\% AA $90.1$\% $\kappa$ $0.84$}}
  \end{subfigure}
  \caption{Sample results from experiment one, Indian Pines dataset. Rows present the three examined architectures, where A9, A3 and A5 corresponds to architectures  \cite{lee2017going}, \cite{yu2017convolutional} and \cite{liu2018siamese} respectively. Columns present the three cases of number of true training samples per class in fine-tuning (5s, 15s and 50s). For each result, the Overall Accuracy (OA), Average Accuracy (AA) and $\kappa$ coefficient are reported. Isolated grey points mark locations of the training samples, and are excluded from the evaluation.}
  \label{fig:exp1ip}
\end{figure*}

\begin{figure*}[!h]
  \centering
  \begin{subfigure}[t]{0.32\textwidth}
    \centering
    \includegraphics[width=0.8\linewidth]{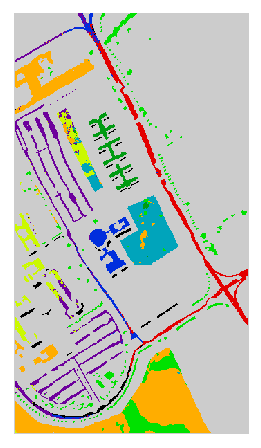}
    \caption{{\footnotesize 5s/A9 OA $79.7$\% AA $88.4$\% $\kappa$ $0.75$}}
  \end{subfigure}\hspace*{0.02\textwidth}
  \begin{subfigure}[t]{0.32\textwidth}
    \centering
    \includegraphics[width=0.8\linewidth]{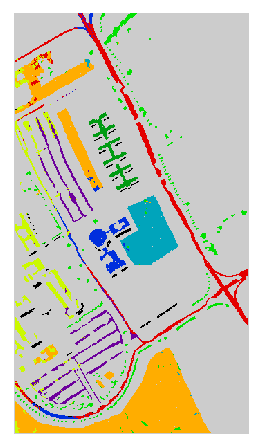}
    \caption{{\footnotesize 15s/A9 OA $91.3.3$\% AA $93.8$\% $\kappa$ $0.89$}}
  \end{subfigure}\hspace*{0.02\textwidth}
  \begin{subfigure}[t]{0.32\textwidth}
    \centering
    \includegraphics[width=0.8\linewidth]{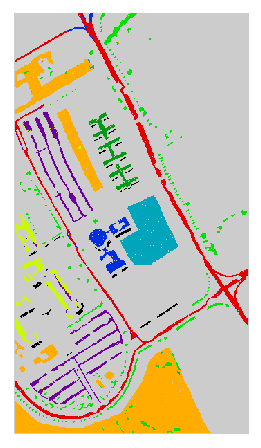}
    \caption{{\footnotesize 50s/A9 OA $97.8$\% AA $98.2$\% $\kappa$ $0.97$}}
  \end{subfigure}\\\vspace{2mm}
  \begin{subfigure}[t]{0.32\textwidth}
    \centering
    \includegraphics[width=0.8\linewidth]{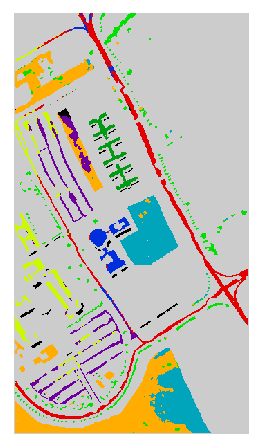}
    \caption{{\footnotesize 5s/A3 OA $81.7$\% AA $91.0$\% $\kappa$ $0.77$}}
  \end{subfigure}\hspace*{0.02\textwidth}
  \begin{subfigure}[t]{0.32\textwidth}
    \centering
    \includegraphics[width=0.8\linewidth]{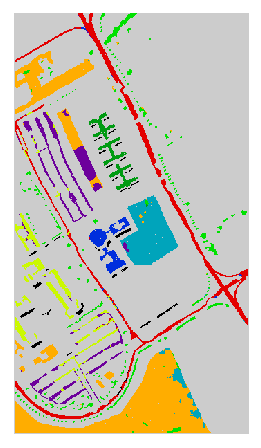}
    \caption{{\footnotesize 15s/A3 OA $92.3$\% AA $94.7$\% $\kappa$ $0.90$}}
  \end{subfigure}\hspace*{0.02\textwidth}
  \begin{subfigure}[t]{0.32\textwidth}
    \centering
    \includegraphics[width=0.8\linewidth]{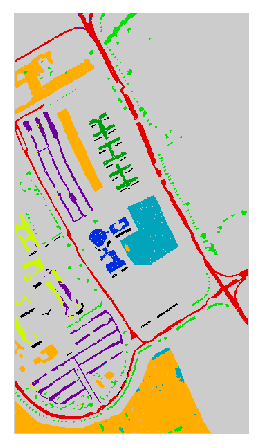}
    \caption{{\footnotesize 50s/A3 OA $97.4$\% AA $97.3$\% $\kappa$ $0.97$}}
  \end{subfigure}\\\vspace{2mm}
  \begin{subfigure}[t]{0.32\textwidth}
    \centering
    \includegraphics[width=0.8\linewidth]{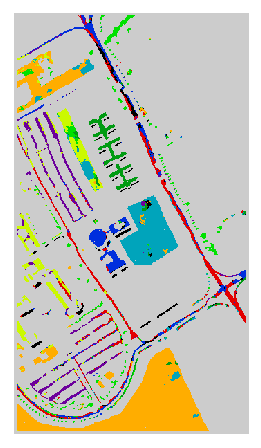}
    \caption{{\footnotesize 5s/A5 OA $77.1$\% AA $79.8$\% $\kappa$ $0.71$}}
  \end{subfigure}\hspace*{0.02\textwidth}
  \begin{subfigure}[t]{0.32\textwidth}
    \centering
    \includegraphics[width=0.8\linewidth]{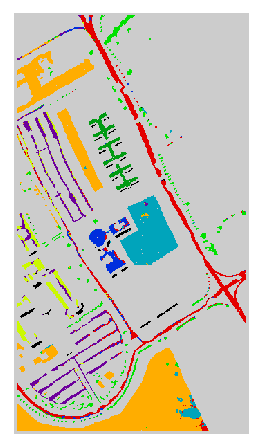}
    \caption{{\footnotesize 15s/A5 OA $90.4$\% AA $88.6$\% $\kappa$ $0.88$}}
  \end{subfigure}\hspace*{0.02\textwidth}
  \begin{subfigure}[t]{0.32\textwidth}
    \centering
    \includegraphics[width=0.8\linewidth]{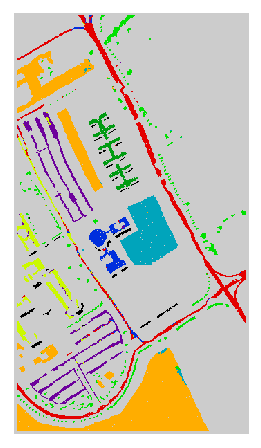}
    \caption{{\footnotesize 50s/A5 OA $96.6$\% AA $96.4$\% $\kappa$ $0.96$}}
  \end{subfigure}
  \caption{Sample results from experiment one, Pavia University dataset. The scheme is identical to the Figure~\ref{fig:exp1ip}.}
  \label{fig:exp1pu}
\end{figure*}

The presented results show that application of the proposed method leads to definite and consistent improvement in accuracy across different images, number of ground truth labels used and network architectures. In all but one case, the improvement is statistically significant, and in some cases approaches 20 percentage points. The most challenging is scenario with 5 training samples per class. Even average overall accuracy achieved by architecture originally examined on small training set~\cite{yu2017convolutional} does not exceed 67\% on Indian Pines dataset. After the application of the proposed method, performance improves up to 72.8\% OA.
The most improvement is seen in the architecture~\cite{lee2017going}, namely on IP dataset with only 5 training samples per class in fine-tuning procedure, it improves from average 52.62 OA to 74.04 OA.  This is to be expected as this architecture has the most potential to benefit from additional training samples. Considering these improvements, it can be summarized that the results of the experiment support stated hypothesis and the validity of the proposed approach. The qualitative evaluation of selected realizations (corresponding to the median score) is presented in Figures~\ref{fig:exp1ip} and~\ref{fig:exp1pu}.

\ctable[
caption = {The result of the first experiment. Each row presents Overall Accuracy (OA), Average Accuracy (AA) and Cohen's kappa ($\kappa$) for given scenario. IP denotes the Indian Pines dataset, PU the Pavia University; further differentiation is for number of samples per class in fine-tuning. Accuracies are given as averages with standard deviations with and without pretraining for the three investigated network architectures.},
label = tab:exp1,
pos = !h
]{cccccccc}{
  \tnote[\dag,\ddag]{Statistically significant improvement, evaluated with Mann--Whitney \emph{U} test, with $P < 0.01$ (\dag) or $P < 0.05$ (\ddag).}
}{ \FL  
&  & \multicolumn{2}{c}{architecture~\cite{lee2017going}} & \multicolumn{2}{c}{architecture~\cite{yu2017convolutional}} & \multicolumn{2}{c}{architecture~\cite{liu2018siamese}}\NN
  
  & & no pretraining&pretraining & no pretraining&pretraining & no pretraining&pretraining 
  \ML
  \multirow{3}{*}{IP 5/class} 
  &OA:& $52.62\pm4.4$ & $74.04\pm4.1^\dag$ &  $66.15\pm4.5$ & $72.80\pm3.2^\dag$ &  $50.05\pm5.1$ & $63.52\pm4.2^\dag$ \\
  &AA:& $58.15\pm2.9$ & $78.83\pm2.6^\dag$ &   $71.42\pm3.9$ & $78.60\pm3.0^\dag$ &  $53.66\pm3.0$ & $65.86\pm3.6^\dag$ \\
  &$\kappa$:&$0.45\pm0.04$ & $0.69\pm0.05^\dag$ &  $0.60\pm0.05$ & $0.68\pm0.04^\dag$ &  $0.41\pm0.05$ & $0.57\pm0.05^\dag$ \ML
        
  \multirow{3}{*}{IP 15/class} 
  &OA:& $67.58\pm3.2$ & $87.04\pm2.4^\dag$ &   $82.61\pm2.8$ & $87.04\pm2.1^\dag$ &  $64.18\pm2.8$ & $75.30\pm1.7^\dag$ \\
  &AA:& $73.82\pm2.7$ & $90.41\pm1.5^\dag$ &  $87.07\pm1.8$ & $90.97\pm1.5^\dag$ &  $67.54\pm2.5$ & $78.40\pm2.1^\dag$ \\
  &$\kappa$:& $0.62\pm0.03$ & $0.85\pm0.03^\dag$ &   $0.79\pm0.03$ & $0.85\pm0.02^\dag$ &  $0.58\pm0.03$ & $0.71\pm0.02^\dag$ \ML

  \multirow{3}{*}{IP 50/class} 
  &OA:& $80.51\pm4.8$ & $93.66\pm1.3^\dag$ &  $93.75\pm1.2$ & $94.65\pm1.0$ &  $81.39\pm1.1$ & $87.06\pm0.9^\dag$ \\
  &AA:& $87.48\pm2.6$ & $95.81\pm0.8^\dag$ &   $95.86\pm0.7$ & $96.62\pm0.8^\dag$ &  $85.10\pm0.9$ & $90.38\pm1.1^\dag$ \\
  &$\kappa$:& $0.77\pm0.05$ & $0.92\pm0.02^\dag$ &   $0.92\pm0.01$ & $0.94\pm0.01$ &   $0.78\pm0.01$ & $0.85\pm0.01^\dag$ \ML

  \multirow{3}{*}{PU 5/class} 
  &OA:& $67.47\pm6.5$ & $80.08\pm7.0^\dag$ &   $73.31\pm4.1$ & $80.33\pm5.2^\dag$ &   $65.55\pm3.8$ & $74.34\pm7.0^\dag$ \\
  &AA:& $76.56\pm3.1$ & $87.66\pm2.9^\dag$ &   $84.67\pm2.6$ & $88.86\pm3.2^\dag$ &  $64.39\pm2.4$ & $76.92\pm3.7^\dag$ \\
  &$\kappa$:& $0.60\pm0.07$ & $0.75\pm0.08^\dag$ &   $0.67\pm0.05$ & $0.76\pm0.06^\dag$ &  $0.56\pm0.04$ & $0.68\pm0.08^\dag$ \ML

  \multirow{3}{*}{PU 15/class} 
  &OA:& $83.63\pm2.7$ & $91.87\pm3.3^\dag$ &   $88.21\pm2.9$ & $91.96\pm2.6^\dag$ &  $75.50\pm2.4$ & $89.33\pm3.4^\dag$ \\
  &AA:& $89.48\pm1.1$ & $94.65\pm1.0^\dag$ &   $93.40\pm1.1$ & $95.01\pm0.8^\dag$ &   $77.59\pm1.2$ & $89.95\pm1.8^\dag$ \\
  &$\kappa$:& $0.79\pm0.03$ & $0.90\pm0.04^\dag$ &   $0.85\pm0.04$ & $0.90\pm0.03^\dag$ &   $0.69\pm0.03$ & $0.86\pm0.04^\dag$ \ML
  
  \multirow{3}{*}{PU 50/class} 
  &OA:& $93.40\pm1.4$ & $97.86\pm0.5^\dag$ &   $96.08\pm0.9$ & $96.84\pm1.2^\ddag$ &   $87.79\pm1.7$ & $96.55\pm0.5^\dag$ \\
  &AA:& $95.47\pm0.7$ & $98.13\pm0.3^\dag$ &   $97.09\pm0.5$ & $97.90\pm0.4^\dag$ &   $89.05\pm0.9$ & $96.37\pm0.3^\dag$ \\
  &$\kappa$:& $0.91\pm0.02$ & $0.97\pm0.01^\dag$ &   $0.95\pm0.01$ & $0.96\pm0.02^\ddag$ &  $0.84\pm0.02$ & $0.95\pm0.01^\dag$ \LL
}

\subsection{Experiment 2}

The results of the experiment are presented in Table~\ref{tab:gridnstripes}. For each grid size or the number of stripes, the overall accuracy and the standard deviation are given. These statistics are based on $50$ experiment runs for each artificial labelling scheme.

It can be seen that the the score rises sharply until the number of artificial classes reaches approximately the number of original classes (at $5 \times 5$, note that the original IP ground truth leaves a sizeable portion of background unmarked, which most probably would contribute some additional classes if marked). After that value, there's a declining trend. It can be noted that the scores are higher with smaller patches. It seems viable to form a conclusion that when the original class number is unknown, it is better to overestimate than underestimate their number. In the latter case, it is possible that even a chance guess would provide a satisfactory performance. The stripes do not form as good a training set as rectangular grid, which confirms the initial supposition that artificial classes should be confined to local areas. Some improvement however is still seen, which supports our overall proposition, that general artificial labelling can be used for improving the DLNN performance without precise estimation of the artificial class patch size.

\ctable[
	caption = {The second experiment results. Grid density describes number of rectangular patches which represent artificial labels for pre-training phase. Num of stripes denotes number of vertical stripes which represent artificial labels for pre-training phase. Accuracies are given as Overall Accuracy for learning of the network based on~\cite{lee2017going} with transfer learning on the Indian Pines dataset.},
	label = tab:gridnstripes,
	pos = h
]{cc|cc}{
}{ \FL
Grid density / model & OA & Num of stripes / model & OA \ML
(2x2)                 & $61.88\pm4.5$ & 2 & $58.53\pm4.9$ \NN
(3x3)                 & $64.45\pm4.1$ & 5 & $67.68\pm4.4$ \NN
(5x5)                 & $75.05\pm4.3$ & 9 & $68.58\pm3.9$ \NN
(7x7)                 & $72.33\pm4.4$ & 16 & $69.25\pm3.7$ \NN
(9x9)                 & $74.06\pm3.6$ & 25& $69.09\pm3.6$ \NN
(14x14)               & $74.13\pm3.7$ & 36 & $69.23\pm3.3$ \NN
(19x19)               & $73.24\pm3.9$ & 49 & $70.08\pm4.0$ \NN
(24x24)               & $73.43\pm4.0$ & 81 & $68.41\pm4.3$ \NN
(29x29)               & $70.19\pm4.6$ &&\NN
(36x36)               & $69.88\pm4.3$ &&\NN
(39x39)               & $68.69\pm4.4$ &&\NN
(48x48)               & $69.25\pm3.2$ &&\NN
(72x72)               & $66.70\pm3.3$ &&\LL
}

\subsection{Experiment 3}

Table \ref{tab:pigments} presents the results of the third experiment. The overall accuracy was calculated based on $n=50$ runs for each examined scenario.

Here, the original performance (GT) can be significantly improved by the grid-based artificial labelling (see results for $5\times 5$, $20\times 20$, $30\times 30$). However, in this case the performance gain can be confronted with a label dataset created from ground truth data (GT-2, GT-10). As can be expected, the ground truth data provides a higher performance; however, the artificial labelling provides half of that gain with no prior information needed. The ground truth experiments GT-2 and GT-10 also confirm the observation that classes split is a better option than joining. The latter observation provides an additional support to the conclusion that more small classes (dense grid) is preferable than few large ones (sparse grid).

\ctable[
	caption = {The third experiment results. Evaluation of pretraining on Pigments dataset using the proposed approach and classes created from ground truth. The objective was to collate the performance of artificial labels of different sizes with those created through splitting or joining the ground truth.},
	label = tab:pigments,
	pos = !h
]{cc|ccc|cc}{
\tnote[a]{No pretraining.}
\tnote[b]{Pretraining with artificial classes (proposed method).}
\tnote[c]{Pretraining with modified ground truth classes (verification).}
}{ \FL
Experiment setting & GT\tmark &  (5x5)\tmark[b]   & (20x20)\tmark[b] & (30x30)\tmark[b] & GT-2\tmark[c] & GT-10\tmark[c] \ML
OA                               & 61.15\% & 68.35\% & 75.70\% & 73.99\% &  75.70\%          & 85.40\%        \LL
}

\subsection{Experiment 4}

\begin{figure}[!h]
  \centering
  \begin{subfigure}[t]{0.49\textwidth}
    \centering
    \includegraphics[width=\linewidth]{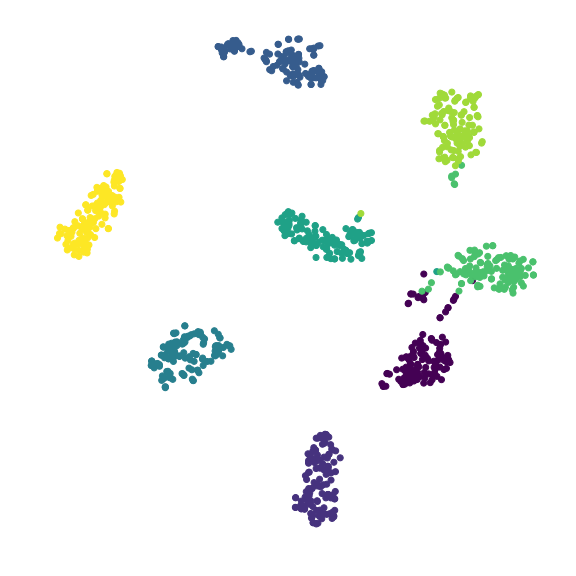}
    \caption{{\footnotesize Activations from network trained using $200$ samples/class}}
  \end{subfigure}\hspace*{0.02\textwidth}
  \begin{subfigure}[t]{0.49\textwidth}
    \centering
    \includegraphics[width=\linewidth]{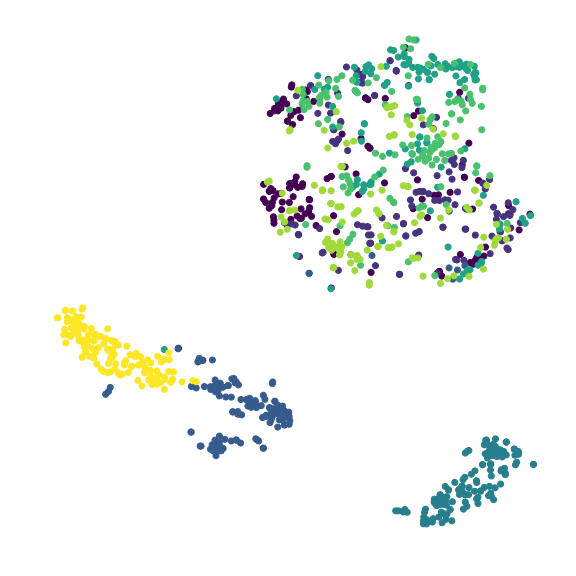}
    \caption{{\footnotesize Activations from network trained using $5$ samples/class}}
  \end{subfigure}\\\vspace{2mm}
  \begin{subfigure}[t]{0.49\textwidth}
    \centering
    \includegraphics[width=\linewidth]{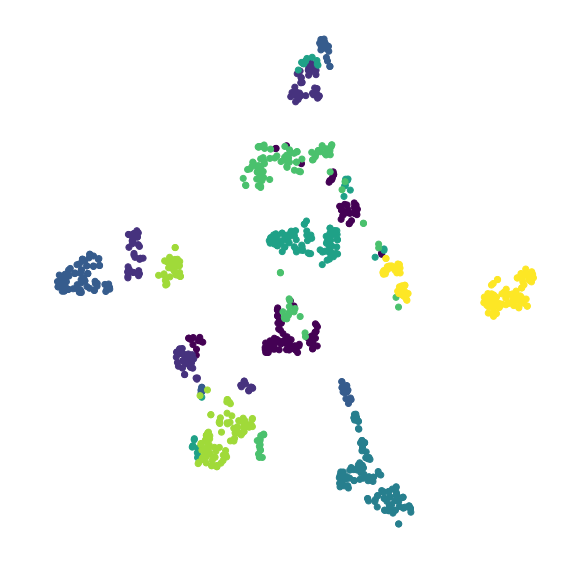}
    \caption{{\footnotesize Activations from network trained using artificial labels only}}
  \end{subfigure}\hspace*{0.02\textwidth}
  \begin{subfigure}[t]{0.49\textwidth}
    \centering
    \includegraphics[width=\linewidth]{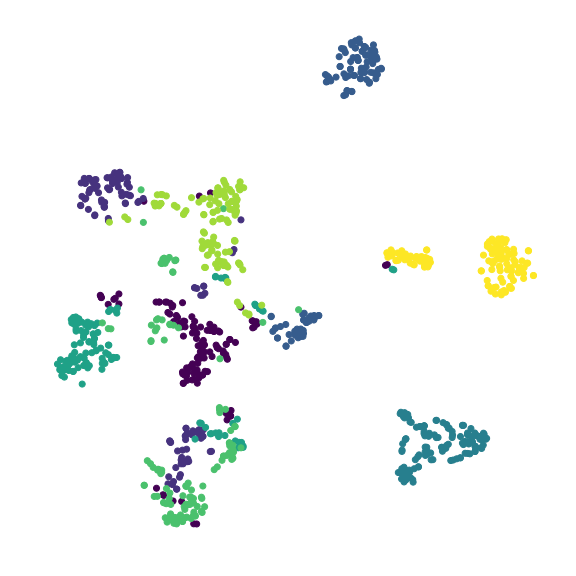}
    \caption{{\footnotesize Activations from network trained with fine-tuning (artificial followed by training labels)}}
  \end{subfigure}\hspace*{0.02\textwidth}

  \caption{The visualisation of the learned parameters for four networks introduced in subsection~\ref{tsne-desc}. Each point represents given sample's activations transformed to the 2-dimensional space using t-SNE algorithm. Different colors represent different classes present on the image.}
  \label{fig:tsne}
\end{figure}

The results of the experiment are presented in Figure~\ref{fig:tsne}. As expected, the network trained using $1600$ true-labeled samples generated good internal representations, which can be seen by the good separability of the classes. In contrast, neural network trained using only $5$ samples per class did not generate representations allowing the separation of samples of different classes. In the case of scenario~\ref{tsne-art}, we can clearly see that the classes were better separated when compared with scenario~\ref{tsne-few}, though of course not as good as in scenario~\ref{tsne-many}. Moreover, the authors did not observe any noticeable differences between scenarios~\ref{tsne-art} and~\ref{tsne-ft}. 

We argue that the presented results provide some suggestion that during neural network training using proposed artificial labelling scheme there is an emergence of useful data-related representations even before the fine-tuning step.

\section{Discussion}

Our results confirm the validity of our proposition: a simple artificial labelling through grouping of the samples based on a local neighbourhood provides an efficient transfer learning scheme. It brings significant improvements of accuracy across datasets and DLNN configurations. The results for different datasets, which have distinctive ground truth layouts suggest that it is not the random alignment with the regularity of a particular ground truth pattern. It is also seen that the local structure is important, as seen in the advantage of grid division over stripe division. The generally better performance of higher over lower number of artificial classes suggests an explanation in that for transfer learning, it is not as important to locate the exact number of classes, but to isolate and learn their components, perhaps for better internal feature representation.

We view the main advantage of the proposed method as enhancing the training of a neural network for hyperspectral remote sensing classification. The proposed pre-training offers a number of benefits:
\begin{enumerate}
	\item Enhance the training of neural networks in hyperspectral classification scenario. With low number of training samples in typical scenarios (e.g. $5-15$/class, sometimes even less) the number of network free parameters can be several orders of magnitude higher than the training data, which poses a risk of overtraining. 
	\item Through splitting the training into two phases, it can be used to shift some of the computational burden of network training to the time before an expert is called in to perform labelling, and make more effective use of his or hers time.
	\item Larger number of training samples available can be of use in case different network architectures are compared for the same problem, or during the searching the hyperparameter space.
\end{enumerate}

\begin{figure*}[!t]
	\centering
	\begin{subfigure}[t]{0.32\textwidth}
		\centering
		\includegraphics[width=\linewidth]{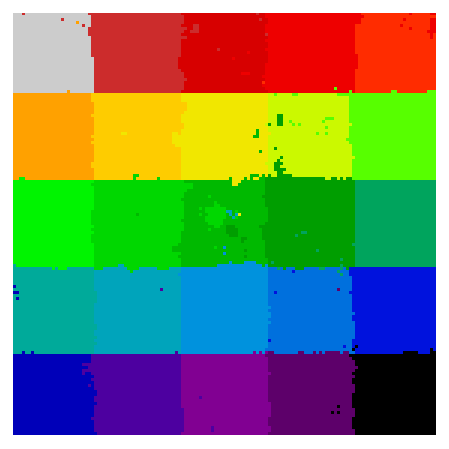}
		\caption{{\footnotesize dataset: IP/architecture: \cite{lee2017going}}}
	\end{subfigure}\hspace*{0.02\textwidth}
	\begin{subfigure}[t]{0.32\textwidth}
		\centering
		\includegraphics[width=\linewidth]{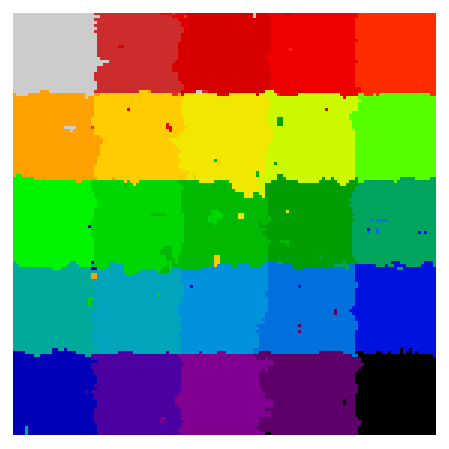}
		\caption{{\footnotesize dataset: IP/architecture: \cite{yu2017convolutional}}}
	\end{subfigure}\hspace*{0.02\textwidth}
	\begin{subfigure}[t]{0.32\textwidth}
		\centering
		\includegraphics[width=\linewidth]{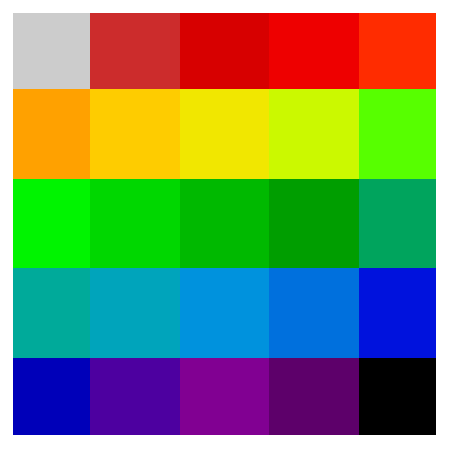}
		\caption{{\footnotesize dataset: IP/architecture: \cite{liu2018siamese}}}
	\end{subfigure}\\
	\vspace{0.01\textheight}
	\begin{subfigure}[t]{0.32\textwidth}
		\centering
		\includegraphics[width=0.8\linewidth]{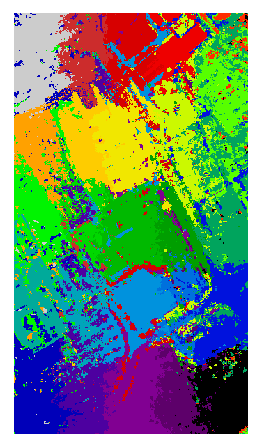}
		\caption{{\footnotesize dataset: PU/architecture: \cite{lee2017going}}}
	\end{subfigure}\hspace*{0.02\textwidth}
	\begin{subfigure}[t]{0.32\textwidth}
		\centering
		\includegraphics[width=0.8\linewidth]{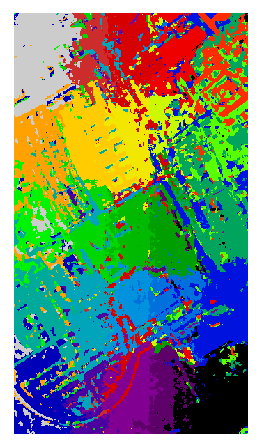}
		\caption{{\footnotesize dataset: PU/architecture: \cite{yu2017convolutional}}}
	\end{subfigure}\hspace*{0.02\textwidth}
	\begin{subfigure}[t]{0.32\textwidth}
		\centering
		\includegraphics[width=0.8\linewidth]{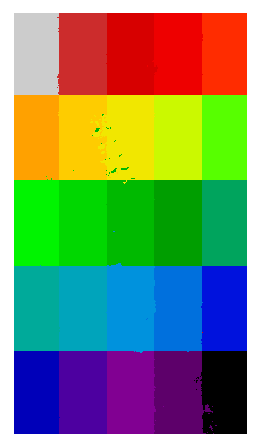}
		\caption{{\footnotesize dataset: PU/architecture: \cite{liu2018siamese}}}
	\end{subfigure}
	\caption{Sample pre-training results. Top row Indian Pines, bottom row Pavia University datasets. Columns present the three architectures studied (based on the works~\cite{lee2017going,yu2017convolutional,liu2018siamese}). Some class structure is visible depending on the dataset and network selected.}
	\label{fig:pretr}
\end{figure*}

An open question is whether a clustering algorithm, like~\cite{wu2018finetuning} or outlier segmentation~\cite{Du2013unsuptl} could be adapted here leading to greater efficiency.
It is probable that a more complex artificial labelling algorithm could outperform the proposed solution; however even in that case, a simple, generally applicable heuristic that improves performance can be of value. Our approach has common motivation with self-taught learning~\cite{Raina2007self}, where we want the classifier to derive high-level input representation from the unlabelled data; however we use the same data for both training stages and instead change the label set. It also avoids combining neural and non-neural approaches, and prevents introducing additional assumptions through the manual selection of the latter.

A qualitative examination of the pre-training results shows that some class structure is visible after pre-training (see examples in Figure~\ref{fig:pretr}). No identifiable features of this structure have been noticed when investigating pre-training images when associated with better or worse final (after fine-tuning) results. However, the general level of structure visible after pre-training relates to the final performance. The network architecture based on the work~\cite{liu2018siamese} is best in learning the artificial classes grid and also the worst at the final classification. The other two networks based on the works~\cite{lee2017going,yu2017convolutional} have more complex pre-training results and correspondingly better final results. This suggests that the training scheme and/or network architecture functions as a form of regularization that prevents overtraining, and that the pre-training classification result can be possibly used to control pre-training and avoid overtraining too. The emergence of partial class structure in the pre-training phase -- which does not use ground truth, hence can be viewed as unsupervised processing -- also suggests that this approach can be adapted to solve unsupervised tasks, e.g. clustering or anomaly detection.

To provide additional verification, we've analysed per-class classification scores for both datasets, using the data from experiment one, and the same Mann--Whitney \emph{U} with $P < 0.05$. As could be expected, performance gains are unequal, as classes differ with their overlap and general difficulty of classification. However, the individual classes showed improvement in most of the cases. Across $198$ tests\footnote{Ten classes for Indian Pines, 11 for Pavia University, each repeated across 9 pairs of network architecture and sample per class number.}, in $104$ cases the improvement was statistically significant; for the remaining cases, in $39$ cases the accuracy of $100$\% was achieved irrespective of pre-training, in $32$ cases pre-training improved the mean of the class score. In the remaining cases where pre-training score mean was lower that the reference, the average difference was below two percentage points. The proposed method thus can be viewed as `not damaging' to individual class scores. 

\begin{figure*}[!t]
	\centering
  \includegraphics[width=0.7\linewidth]{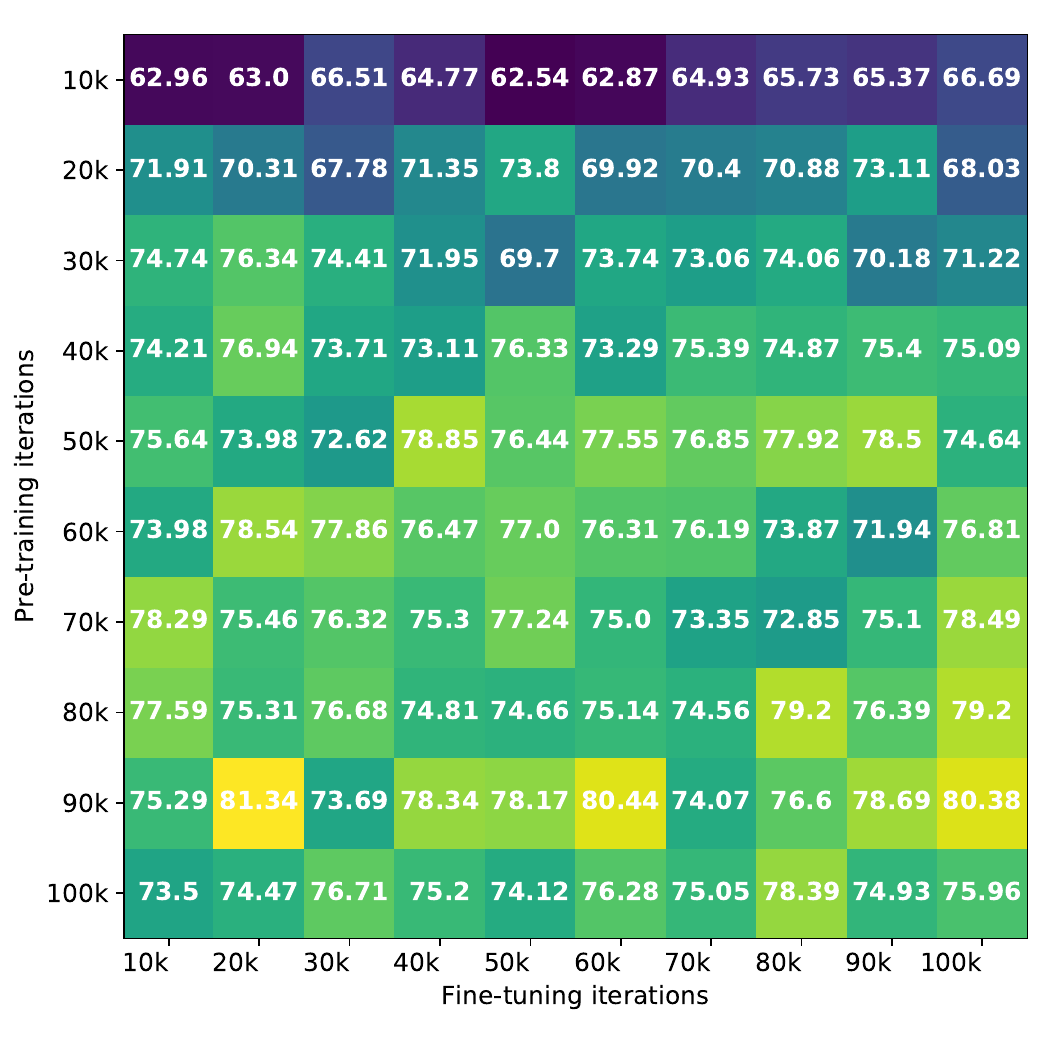}
	\caption{The classification accuracies of the networks trained on $5$ samples/class and tested with the rest of the image using the Indian Pines dataset and neural network based on~\cite{lee2017going}. On the y-axis, the number of epochs for the pre-training stage is written, while on the x-axis the number of epochs for the fine-tuning stage is written. The resuts suggest the relative importance of the pre-training stage in comparison to the fine-tuning stage.}
	\label{fig:time}
\end{figure*}

Additionally, a batch of experiments were performed for sensitivity analysis of small variations of hyperparameter setting; the results were very similar to those presented. A separate experiment was conducted analysing time-requirements when training the networks. The results of the experiments are presented in Figure~\ref{fig:time}. The results show that it is more important to train the network during pre-training stage than during the fine-tuning stage (one can clearly see the results getting better when moving vertically within a grid from Figure~\ref{fig:time}, as opposed to moving horizontally). As one can see, in the case of the lower number of pre-training iterations (10k-50k), even moderate increase leads to definite improvement in the accuracy of the classification. The results also suggest that it could be possible to reduce the time of training in both of the stages without sacrificing the effectiveness of classification. Moreover, it can be presumed that choosing a different number of iterations of the pre-training and fine-tuning stages could lead to achieving even better results than the ones presented in this work (for example, when training the network for 90k iterations in the pre-training stage, and 20k iterations in the fine-tuning stage, it was possible to achieve the accuracy of 81.34\%).

Analysing the results from the Table~\ref{tab:exp1}, one can notice that pre-training improves accuracy in some networks more than in others. We suspect that an important factor determining such differences is the capacity of neural networks. We argue this with fact that artificial neural network with greater number of parameters is able to better process information contained in the entire image, which we utilize in the pre-training phase.
Therefore, the architecture~\cite{yu2017convolutional} with the smallest number of parameters achieves a smaller increment of the accuracy in comparison to other two networks.

However, one must to be aware that there are a number of other factors that affect network performance. In particular, architectures~\cite{lee2017going,yu2017convolutional} were designed for the task of HSI classification. With an emphasis on the architecture~\cite{yu2017convolutional}, which has been studied on a small training data sets, and therefore has competitive accuracy even without pre-training. On the other hand, architecture~\cite{liu2018siamese} was designed for a slightly different training regimen, which may explain the fact that it achieves worse results than the other two.

Our approach could be used for semi-automatic systems like~\cite{Girshick2014rich}, which use only a part of the annotation, and could be made fully unsupervised. Furthermore, we believe this is one approach for self-taught learning~\cite{Raina2007self}, that can be helpful in diverse application of deep learning models. We note, however, that optimization would require further studies to address the issue of which layers benefit most of this scheme, i.e. similar to~\cite{yosinski2014transfer}. Our experiments show that the proposed scheme is largely resistant to the incorrect estimation of the number of classes, hence its parametrization can be considered low-cost. It can be also viewed as a confirmation of traditional software development principle of `divide and conquer', as of even older proverb, \emph{`divide et impera'}.

\section{Conclusions}

We have presented and verified a simple method pre-training of DLNN for hyperspectral classification based on the hypothesis that spatial similarity of unlabelled data points can be utilized to gain accuracy in hyperspectral classification.
In the first experiment, we showed that for all three neural network architectures tested, and for the all two reference datasets, the proposed procedure leads to an improvement of classification efficiency for small number of training samples.
In the second and third experiments, we analysed the properties of proposed method; the obtained results suggest that the number and shape of the pixel blobs have an impact on the effectiveness of the method. Specifically, we conclude from the second experiment that it is safer to underestimate the size of a label cluster rather than overestimate and simultaneously reduce chance of joining separate classes. This conclusion is in line with results of the third experiment, from which we also conclude that it is better to split ground truth classes than join them.

The absence of training labels requirement provides an important advantage: it shifts the need of expert's participation and data labelling from the start of the data analysis process to its late stages. This allows for the use of the potentially long time from the acquisition to the start of data interpretation stage for pre-training the network, and decreases the delay between expert's labelling to getting the classification result. Considering the length of time required to train deep neural networks, this is a significant advantage for their applications. An additional benefit is that multiple unannotated images can be used in the pre-training stage, potentially increasing the robustness of the result.

\section*{Acknowledgements}
This work has been partially supported by the projects: `Representation of dynamic 3D scenes using the Atomic Shapes Network model' financed by the National Science Centre, decision DEC-2011/03/D/ST6/03753 and `Application of transfer learning methods in the problem of hyperspectral images classification using convolutional neural networks' funded from the Polish budget funds for science in the years 2018-2022, as a scientific project under the ,,Diamond Grant'' program, no. DI2017 013847. M.O. acknowledges
support from Polish National Science Center scholarship 2018/28/T/ST6/00429.

This research was supported in part by PLGrid Infrastructure. The authors would like to thank Laboratory of Analysis and Nondestructive Investigation of Heritage Objects (LANBOZ) in National Museum in Kraków for providing the pigments dataset, in particular to Janna Simone Mostert for her help in the preparation of paintings and Agata Mendys for acquisition of the dataset. The authors also thank Zbigniew Puchała for help in carrying out statistical analysis of the results. Additionally authors thank Yu et al~\cite{yu2017convolutional} for sharing the code.

\bibliographystyle{unsrt}
\bibliography{transfer_learning}

\end{document}